\documentclass{article}
\usepackage{ijcai20}

\usepackage{times}
\usepackage{soul}
\usepackage{url}
\usepackage[utf8]{inputenc}
\usepackage[small]{caption}
\usepackage{graphicx}
\usepackage{amsmath}
\usepackage{amsthm}
\usepackage{booktabs}
\usepackage{algorithm}
\usepackage{algorithmic}
\usepackage{multirow}
\usepackage{color}

\usepackage{pifont} 
\newcommand{\cmark}{\ding{51}}%
\newcommand{\xmark}{\ding{55}}%

\newcommand{\dataname}{\textsc{L}ogiQA}
\newcommand{\datanamecn}{\textsc{C}hinese \textsc{L}ogiQA}

\title{LogiQA: A Challenge Dataset for Machine Reading Comprehension \\with Logical Reasoning}

\author{
Jian Liu$^{1*}$
\and
Leyang Cui$^{2,3*}$
\and
Hanmeng Liu$^{2,3}$
\and
Dandan Huang$^{2,3}$
\and
Yile Wang$^{2,3}$\and
Yue Zhang$^{2,3}$$^\dagger$
\affiliations
$^1$School of Computer Science, Fudan University\\
$^2$School of Engineering, Westlake University\\
$^3$Institute of Advanced Technology, Westlake Institute for Advanced Study\\
\emails
jianliu17@fudan.edu.cn,
\{cuileyang, liuhanmeng, huangdandan, wangyile, zhangyue\}@westlake.edu.cn,
}

\begin{document}

\maketitle
\renewcommand{\thefootnote}{\fnsymbol{footnote}}
\footnotetext[1]{Equal contribution.}
\footnotetext[2]{Corresponding Author}

\begin{abstract}
Machine reading is a fundamental task for testing the capability of natural language understanding, which is closely related to human cognition in many aspects. With the rising of deep learning techniques, algorithmic models rival human performances on simple QA, and thus increasingly challenging machine reading datasets have been proposed. Though various challenges such as evidence integration and commonsense knowledge have been integrated, one of the fundamental capabilities in human reading, namely logical reasoning, is not fully investigated. We build a comprehensive dataset, named {\bf \dataname}, which is sourced from expert-written questions for testing human {\bf Logi}cal reasoning. It consists of 8,678 QA instances, covering multiple types of deductive reasoning. Results show that state-of-the-art neural models perform by far worse than human ceiling. Our dataset can also serve as a benchmark for re-investigating logical AI under the deep learning NLP setting. The dataset is freely available at ~\url{https://github.com/lgw863/LogiQA-dataset}.

\end{abstract}

\section{Introduction}
Machine reading~\cite{hermann2015teaching,sar} is a popular task in NLP, which is useful for downstream tasks such as open domain question answering and information retrieval. In a typical task setting, a system is given a passage and a question, and asked to select a most appropriate answer from a list of candidate answers. With recent advances of deep learning in NLP, reading comprehension research has seen rapid advances, with a development from simple factual question answering~\cite{squad} to questions that involve the integration of different pieces of evidences via multi-hop reasoning~\cite{constructing,hotpotqa} and questions that involve commonsense knowledge outside the given passage~\cite{ostermann,cosmos}, where more varieties of challenges in human reading comprehension are investigated.

\begin{figure}[t!]
\centering
\setlength{\abovecaptionskip}{0.1cm}
\setlength{\belowcaptionskip}{0.2cm}
\includegraphics[width=0.48\textwidth]{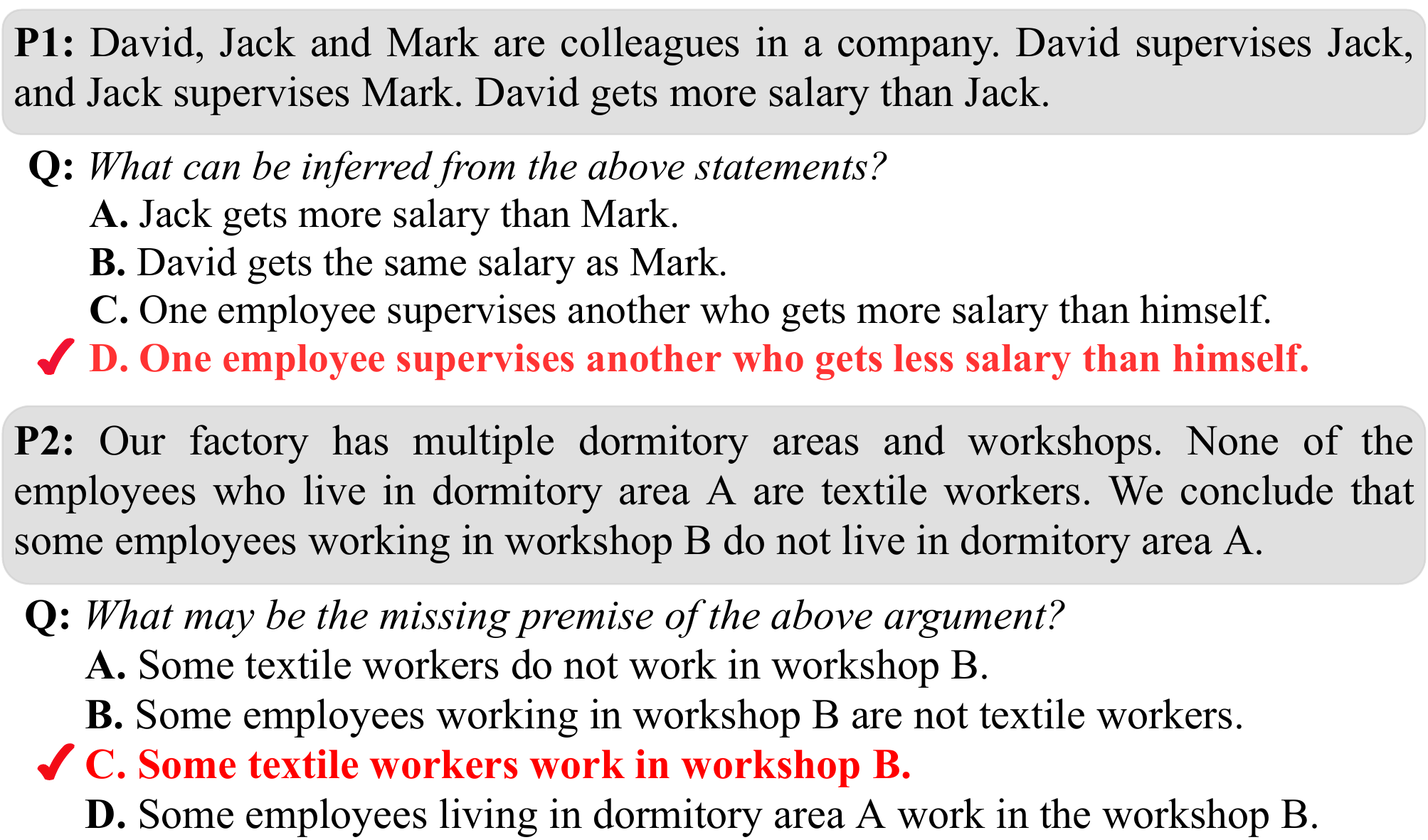}
\caption{Examples of \dataname. 
({\color{red} {\cmark}} indicates the correct answer.) }
\label{fig1:example}
\vspace{-0.5cm}
\end{figure}

One important aspect of human reading comprehension and question answering is logical reasoning, which was also one of the main research topics of early AI~\cite{mccarthy1989artificial,colmerauer1996birth}. To this end, there has been relatively very few relevant datasets in modern NLP. Figure \ref{fig1:example} gives two examples of such problems. In particular, P1 consists of a paragraph of facts, and a question that asks the testee to select a valid conclusion by taking the facts as premises. In order to select the correct candidate, a machine is expected to understand the premises and the candidate answers. The correct answer can be found by categorical reasoning. P2 of Figure \ref{fig1:example} is more challenging in providing a premise and a conclusion in the paragraph, while asking for a missing premise. In particular, three sets of workers are involved, including those living in dormitory area A, those who work in workshop B and those who are textile workers. A testee can find the answer by drawing logical correlations between the three sets of workers. 

The type of machine reading comprehension questions above requires a combination of natural language understanding and logical reasoning. Compared with factual question answering, lexical overlap between the paragraph and the candidate answers plays a relatively less important role. Compared with commonsense reading comprehension, such questions do not rely heavily on external knowledge. Instead they focus on logical inference. In this aspect, the questions can be viewed as a re-investigation of logical AI \cite{mccarthy1989artificial,nilsson1991logic} in the age of neural NLP. One solution can be to perform semantic parsing into formal logic representation, and then perform symbolic reasoning. However, with the abundance of deep learning methods in the NLP toolkit, it can also be interesting to learn the potentials of neural AI for solving such tasks.

To facilitate such research, we create a new reading comprehension dataset, \dataname, which contains 8,678 paragraph-question pairs, each with four candidate answers. Our dataset is sourced from publically available logical examination papers for reading comprehension, which are designed by domain experts for evaluating the logical reasoning ability and test participants. Thus the quality and topic coverage of the questions are reliable. We manually select problems from the original dataset, filtering out problems that involve figures, charts, or those that are heavy of mathematics, and ensuring a wide coverage of logical reasoning types, including categorical reasoning, conditional reasoning, disjunctive reasoning and conjunctive reasoning~\cite{concise-logic}.

To establish baseline performances on \dataname, we explore several state-of-the-art neural models developed for reading comprehension. Experimental results demonstrate a significant gap between machine (35.31\% accuracy) and human ceiling performance (96.00\%). We provide detailed analysis to give insights into potentially promising research directions. 

\section{Related Work}

\paragraph{Existing Datasets For Reading Comprehension}

A seminal dataset for large-scale reading comprehension is SQuAD \cite{squad}, which requires selecting a factual answer from all possible spans in a given passage. Many neural methods have been developed for this dataset, achieving results that rival human testees. As a consequence, more reading comprehension datasets with increasing challenges are proposed. These datasets can be classified according to the main challenges. In particular, TriviaQA \cite{triviaqa} requires evidence integration across multiple supporting documents to answer the questions. DuoRC \cite{duorc} and Narrative QA \cite{narrativeqa} raise challenges by introducing two passages about the same facts. \citeauthor{constructing}~\shortcite{constructing} and HotpotQA \cite{hotpotqa}  test models for text understanding with sequential multi-step reasoning. Drop \cite{drop} tests discrete numerical reasoning over the context. MuTual \cite{mutual} tests dialogue reasoning ability via the next utterance prediction task. These datasets are factual in the sense that the answer (or candidate in multi-choice-questions) is mostly a text span in the given passage. Several types of reasoning are necessary, such as geolocational reasoning and numerical computation. Different from these datasets, our dataset contains answers not directly included in the input passage, and requires comprehensive reasoning methods beyond text matching based techniques.

\begin{table}
\centering
\setlength{\abovecaptionskip}{0.2cm}
\setlength{\belowcaptionskip}{-0.2cm}
\footnotesize
\begin{tabular}{cccc}
\hline
{\bf Dataset}  & {\bf Logic}  & {\bf Domain} & {\bf Expert}\\
\hline
SQuAD & \xmark  & Wikipedia & \xmark     \\
TriviaQA & \xmark  & Trivia & \xmark     \\
RACE & \xmark  & Mid/High School Exams  &\cmark   \\
DuoRC  &\xmark   & Wikipedia \& IMDb  & \xmark     \\
Narrative QA &\xmark   & Movie Scripts, Literature  & \xmark \\
DROP    &\xmark   & Wikipedia   & \xmark     \\
COSMOS     &\xmark   & Webblog   & \xmark     \\
MuTual & \xmark & Daily Dialogue & \cmark \\
\hline
\dataname(Ours)       & \cmark  & Civil Servants Exams  &\cmark \\
\hline
\end{tabular}
\caption{Comparison with existing reading comprehension datasets. ``Logic" indicates that dataset mainly requires logical reasoning. ``Expert" indicates that dataset is designed by domain experts.}
\label{tab:plain}
\end{table}

Similar to our dataset, recent datasets for commonsense reasoning, including MCScript \cite{ostermann} and COSMOS \cite{cosmos}, also contain candidate answers not directly included in the input passage. They test a model's capability of making use of external background knowledge about spatial relations, cause and effect, scientific facts and social conventions. In contrast, our dataset focuses on logical reasoning and most of the necessary facts are directly included in the given passage. In addition, most of the existing datasets are labeled by crowd sourcing. In contrast, our dataset is based on examination problems written by human experts for students, and therefore has a better guarantee of the quality. This is particularly important for datasets that involve abstract reasoning skills.

\begin{figure*}[t!]
\centering
\setlength{\belowcaptionskip}{-0.1cm}
\includegraphics[width=0.99\textwidth]{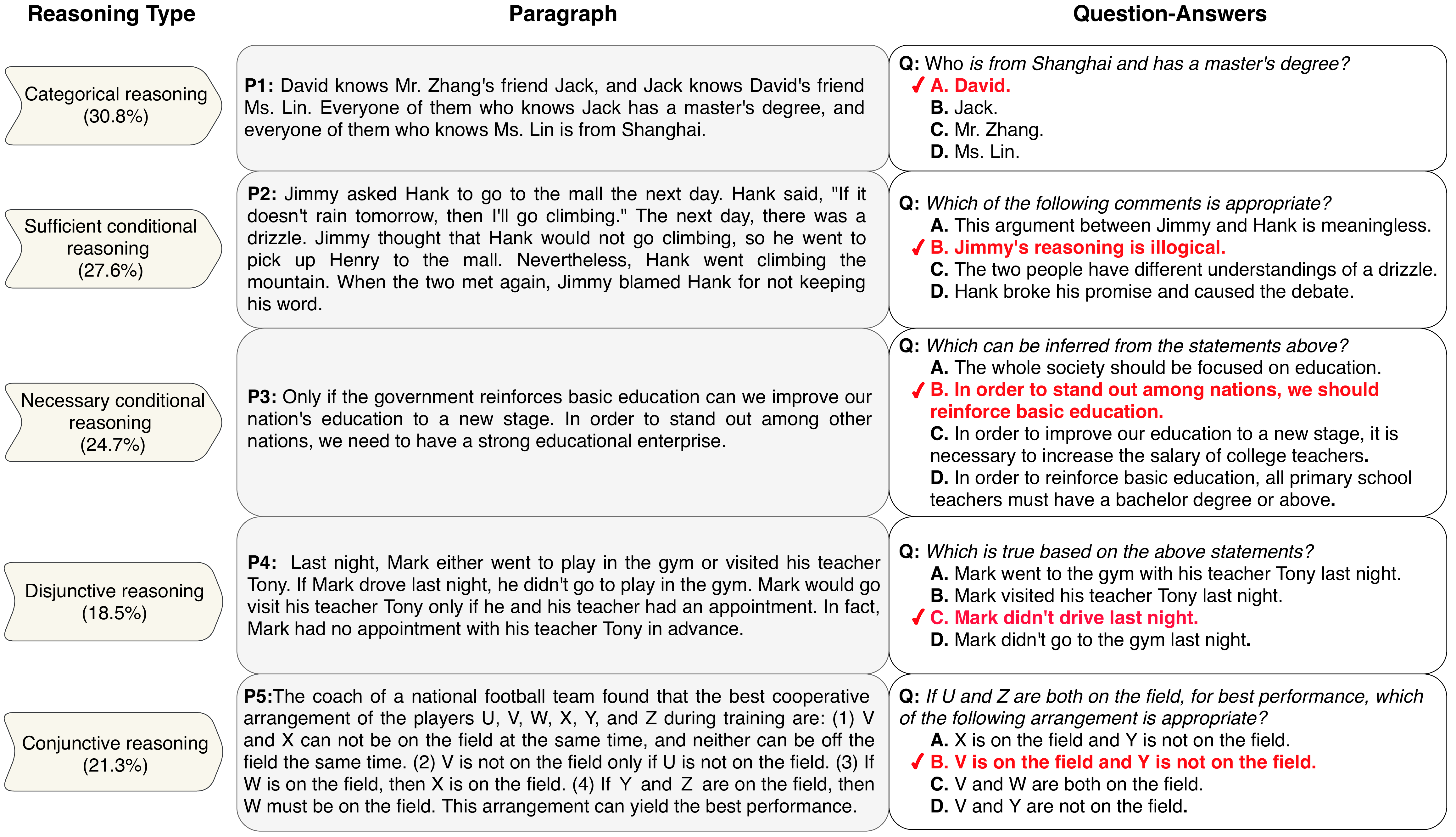}
\caption{Examples of each type of logical reasoning in \dataname. 
({\color{red} {\cmark}} indicates the correct answer.) 
}
\label{fig2:example}
\end{figure*}

The correlation and difference between our dataset and existing QA datasets are shown in Table~\ref{tab:plain}.


\paragraph{Logical Reasoning}

There have been existing datasets related to logical reasoning. In particular, \citeauthor{argument-reasoning}\shortcite{argument-reasoning} design a dataset for argument reasoning, where a claim is given and the model is asked to choose a correct premise from two candidates to support the claim. Similar to our dataset, the dataset concerns deductive reasoning. The biggest difference between our dataset and this dataset is that ours is a machine reading comprehension test while theirs focuses on argumentation. The form of their task is closer to NLI as compared with reading comprehension. In addition, our dataset has more instances (8,678 vs 1,970), more choices per question (4 vs 2) and is written by relevant experts rather than being crowd-sourced. CLUTRR \shortcite{clutrr} is a dataset for inductive reasoning over family relations. The input is a given passage and a query pair, the output is a relationship between the pair. The dataset concerns reasoning on a fixed domain (i.e., family relationship), which is in line with prior work on social relation inference \cite{bramsen2011extracting}. In contrast, our dataset investigates general logical reasoning.

\paragraph{Datasets from Examinations}

Our dataset is also related to datasets extracted from examination papers, aiming at evaluating systems under the same conditions as how humans are evaluated. For example, the AI2 Elementary School Science Questions dataset~\cite{khashabi2016question} contains 1,080 questions for students in elementary schools; RACE~\cite{race} collects passages and questions from the English exams for middle school and high school Chinese students in ages between 12 to 18. These datasets are based on English tests, examining testees on general language understanding. They target students in language learning. In contrast, our datasets are based on examinations of logical skills in addition to language skills. To the best of our knowledge,~\dataname \ is the first large-scale dataset containing different types of logical problems, where problems are created based on exams designed by human experts. 

\section{Dataset}

\subsection{Data Collection and Statistics}

We construct \dataname \ by collecting the logical comprehension problems from publically available questions of the National Civil Servants Examination of China, which are designed to test the civil servant candidates' critical thinking and problem solving. We collected raw data released at the official website, obtaining 13,918 paragraph-question-choice triples with the correct answers. 

The following steps are conducted to clean the raw data. First, we remove all the instances that do not have the format of our problem setting, i.e., a question is removed if the number of candidate choices is not four. Second, we filter all the paragraphs and questions that are not self-contained based on the text information, i.e. we remove the paragraphs and questions containing images or tables. We also remove all questions containing the keywords “underlined” and “sort sentences”, since it can be difficult to reproduce the effect of underlines and sentence number order for a typical machine reader. Finally, we remove all duplicated paragraph-question pairs. The resulting dataset contains 8,678 paragraph-questions pairs.

\begin{table}[t!]
\footnotesize
\centering
\setlength{\belowcaptionskip}{-0.2cm}
\begin{tabular}{l|c}
\toprule 
{\bf Parameter} & {\bf Value} \\ 
\midrule 
\# Paragraphs-Question Pair & \ 8,678 \\
Ave./Max. \# Tokens / Paragraph & 76.87 / 323 \\
Ave./Max. \# Tokens / Question & 12.03 / \ \ 54 \\
Ave./Max. \# Tokens / Candidate Answer & 15.83 / 111 \\
\bottomrule
\end{tabular}
\caption{Statistics of \dataname.}
\label{statistics}
\end{table}

Since the original dataset was written in Chinese, five professional English speakers are employed to translate the dataset manually. To ensure translation quality, we further employ three proofreaders. A translated instance is sent back to the translators for revision if proofreaders reject the instance. The detailed statistics for \dataname \ is summarized in Table \ref{statistics}. Compared with existing reading comprehension datasets, the average paragraph length is relatively small since logical reasoning problems do not heavily rely on complex context.

We also release the Chinese version of \dataname \ (named as \datanamecn) for Chinese reasoning-based reading comprehension research.

\subsection{Reasoning Types of the Dataset}
\label{sec:reasoning type}
The test set of our benchmark consists of 867 paragraph-question pairs. We manually categorize the instances according to the five types of logical reasoning defined by Hurley \shortcite{concise-logic}, including categorical reasoning, sufficient conditional reasoning, necessary conditional reasoning, disjunctive reasoning and conjunctive reasoning. These types of reasoning belong to deductive reasoning, for which a definite conclusion can be derived given a set of premises. As a result, such reasoning can be most suitable for evaluating performances quantitatively. Figure \ref{fig2:example} shows the statistics and representative examples of the reasoning types in our dataset. Note that the sum of percentage values is above 100\%, which is because one problem can involve multiple types of reasoning. Formally, the five types of reasoning can be described as follows:
\begin{itemize}
\item {\bf Categorical reasoning:} The goal is to reason whether a specific concept belongs to a particular category. This type of reasoning is commonly associated with quantifiers such as \emph{``all/everyone/any"}, \emph{``no"} and \emph{``some"}, etc.
\item {\bf Sufficient conditional reasoning:} The type of hypothetical reasoning is based on conditional statements of the form ``\emph{If P, then Q}", in which \emph{P} is the antecedent and \emph{Q} is the consequent. 
\item {\bf Necessary conditional reasoning:} This type of hypothetical reasoning is based on conditional statements of the form ``\emph{P} only if \emph{Q}", ``\emph{Q} whenever \emph{P}", etc., where Q is a necessary condition for P. 
\item {\bf Disjunctive reasoning:} In this type of reasoning, the premises are disjunctive, in the form ``\emph{either . . . or . . .}" , where the conclusion holds as long as one premise holds. 
\item {\bf Conjunctive reasoning:} In this type of reasoning, the premises are conjunctives, in the form ``\emph{both …and…}", where the conclusion holds only if all the premises hold.
\end{itemize}


\section{Methods}
\label{sec:Baseline Methods}

 We evaluate the performances of typical reading comprehension models, including rule-based methods, deep learning methods as well as methods based on pre-trained contextualized embedding. In addition, human performances are evaluated and ceiling performances are reported.

\paragraph{Rule-Based Methods} We adopt two rule-based methods, which rely on simple lexical matching. In particular, {\it word matching} \cite{word-matching} is a baseline that selects the candidate answer that has the highest degree of unigram overlap with the given paragraph-question pair; {\it sliding window} \cite{sliding-window} calculates the matching score for each candidate answer by extracting TF-IDF type features from n-grams in the given paragraph-question pair. 

\paragraph{Deep Learning Methods} Most existing methods \cite{sar,gar,dual-co-matching} find the answer by text matching techniques, calculating the similarity between the given paragraph, the question and each candidate answer. In particular, {\it Stanford attentive reader} \cite{sar} computes the similarity between a paragraph-question pair and a candidate answer using LSTM encoding and a bi-linear attention function; {\it gated attention reader} \cite{gar} adopts a multi-hop architecture with a more fine-grained mechanism for matching candidate answers with paragraph-question pairs; {\it co-matching network} \cite{dual-co-matching} further enhances matching the paragraph-question pair and paragraph-candidate answer pair by encoding each piece of text and calculating matching score between each pair, respectively. 

\paragraph{Pre-trained Methods} Pre-trained models give the current state-of-the-art results on machine reading. Different from the above deep learning methods, pre-trained methods consider the paragraph, the question and each candidate answer as one concatenated sentence, using a pre-trained contextualized embedding model to encode the sentence for calculating its score. Given four candidate answers, four concatenated sentences are constructed by pairing each candidate answer with the paragraph and question, and the one with the highest model score is chosen as the answer. In particular, {\it BERT} \cite{bert} treats the paragraph as sentence A and the concatenation of the question and each candidate as sentence B, before further concatenating them into {\it [CLS] A [SEP] B [SEP]} for encoding; {\it RoBERTa} \cite{liu2019roberta} replaces the BERT model using the RoBERTa model. The hidden state of the {\it[CLS]} token is used for MLP + softmax scoring. The embedding models are fine-tuned during training.

\begin{table*}[t!]
    \centering
    \setlength{\abovecaptionskip}{0.1cm}
    \setlength{\belowcaptionskip}{-0.4cm}
    \scriptsize
    \begin{tabular}{c|c|c|c|c|c}
    \toprule
       \multicolumn{1}{c|}{\multirow{2}{*}{\bf{Category}}} & \multicolumn{1}{c|}{\multirow{2}{*}{\bf{Model}}}  & \multicolumn{2}{c|}{\dataname}& \multicolumn{2}{c}{\datanamecn}\\ \cmidrule{3-6}
    \multicolumn{1}{c|}{}&  \multicolumn{1}{c|}{} & {\bf Dev} & {\bf Test} & {\bf Dev} & {\bf Test} \\
    \midrule
    & Random(theoretical) & 25.00 & 25.00 & 25.00 & 25.00 \\
    \midrule
         \multirow{2}{*}{Rule-based} & Word Matching \cite{word-matching} & 27.49 & 28.37 & 26.55 & 25.74 \\
         & Sliding Window \cite{sliding-window} & 23.58 & 22.51 & 23.85 & 24.27 \\
    \midrule
        \multirow{3}{*}{Deep learning} & Stanford Attentive Reader \cite{sar} & 29.65 & 28.76 & 28.71 & 26.95 \\ 
        & Gated-Attention Reader \cite{gar} & 28.30 & 28.98 & 26.82 & 26.43 \\
        & Co-Matching Network \cite{dual-co-matching} & 33.90 & 31.10 & 30.59 & 31.27 \\
    \midrule
        \multirow{2}{*}{Pre-trained} & BERT \cite{bert} & 33.83 & 32.08 & 30.46 & 34.77 \\
        & RoBERTa \cite{liu2019roberta} & {\bf35.85} & {\bf35.31} & {\bf39.22} & {\bf37.33} \\
    \midrule
    \multirow{2}{*}{Human} & Human Performance & - &86.00& - &88.00 \\
    & Ceiling Performance & - &95.00& - &96.00 \\
    \bottomrule
    \end{tabular}
    \caption{Main results on \dataname\ (accuracy\%).}
    \label{tab:main_experiment}
\end{table*}

\paragraph{Human Performance}
We employ three post-graduate students for human performance evaluation, reporting the average scores on 500 randomly selected instances from the test set. For calculating the ceiling performances, we consider a question as being correctly answered if one of the students gives the correct answer.

\paragraph{Implementation Details}
We re-implement the rule-based methods strictly following the original papers \cite{word-matching,sliding-window}.
For the deep learning methods, we directly use the implementations released in the original papers. 100-dimensional Glove word embeddings are used as embedding initialization. For pre-trained methods, we follow the HuggingFace implementation~\cite{Wolf2019HuggingFacesTS}. We take the off-the-shelf model BERT-base and RoBERTa-base for \dataname, and Chinese BERT-base and Chinese RoBERTa-base ~\cite{cui2019pre} for \datanamecn. All hyper-parameters are decided by the model performance on the development sets.


\section{Results}
We randomly split the dataset, using 80\% for training, 10\% for development and the remaining 10\% for testing. Table~\ref{tab:main_experiment} shows the results of the models discussed in the previous section. In particular, the human performance is 86.00\% and the ceiling performance is 95.00\%, which shows that the difficulty level of the dataset is not high for human testees. In contrast, all of the algorithmic models perform significantly worse than human, demonstrating that the methods are relatively weak in logical reasoning reading comprehension. In addition, results on the Chinese dataset are on the same level compared with those on the English dataset.

In particular, the rule-based methods give accuracies of 28.37\% and 22.51\%, respectively, the latter being even lower than a random guess baseline. This shows that the questions are extremely difficult to solve by using lexical matching alone. Figure~\ref{fig1:example} serves as one intuitive example. The deep learning methods such as the Stanford attentive reader, the gated attention reader and the co-matching network give accuracies around 30\%, which is better compared with the random guess baseline but far behind human performance. One likely reason is that the methods are trained end-to-end, where it turns out difficult for attention-based text matching to learn underlying logical reasoning rules. 

It has been shown that pre-trained models have a certain degree of commonsense and logical capabilities \cite{cosmos}. On \dataname, such models give better performances compared with the methods without contextualized embeddings. However, the best result by RoBERTa is 35.31\%, still much below human performance. This shows that knowledge in pre-trained models is rather weak for logical reasoning. It remains an open question on how deep learning machine readers can be equipped with strong reasoning capability.
\begin{figure}[t!]
\centering
\setlength{\belowcaptionskip}{-0.2cm}
\includegraphics[width=0.44\textwidth]{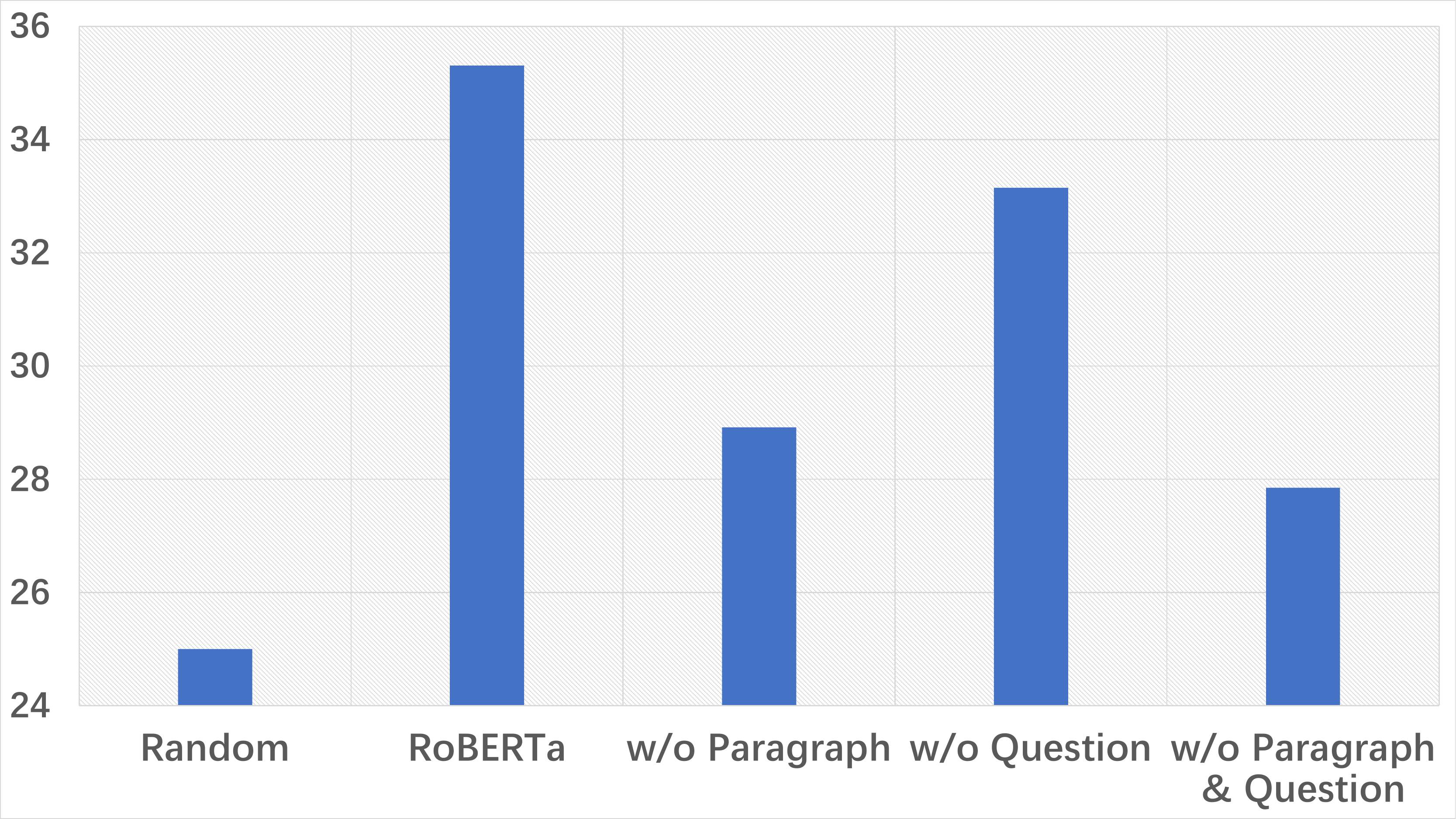}
\caption{Ablation of paragraph or question (accuracy\%).}
\label{fig3:example}
\end{figure}

\section{Discussion}
We give detailed analysis based on the empirical results of RoBERTa and other models on \dataname \ test set.
\subsection{Ablation Tests}

Following recent studies, we conduct a set of ablation experiments using RoBERTa to measure bias in the dataset by checking the performance based on the partial information \cite{cai2017pay}. Figure~\ref{fig3:example} shows the results on the test set. There is a significant drop of accuracies without the paragraph, the question or both, which indicates that the bias on the dataset is weak. In particular, without the input paragraph, the accuracy drops from 35.31\% to 28.92\%, which is comparable to a drop from 67.1\% to 55.9\% by the same model on the COSMOS dataset \cite{cosmos}, and 66.0\% to 52.7\% on the Social IQa dataset \cite{Social-IQa}.

\begin{table}[t!]
    \centering
    \setlength{\abovecaptionskip}{0.1cm}
    \setlength{\belowcaptionskip}{-0.4cm}
    \scriptsize
    \begin{tabular}{c|cc}
        \toprule
         {\bf Model} & {\bf Dev} & {\bf Test} \\
        \midrule
        Random(theoretical) & 25.00 & 25.00 \\
         \midrule
         RoBERTa$_{\text{\dataname}}$ & 35.85 & 35.31 \\
         \midrule
         RoBERTa$_{\text{RACE}}$ & 29.19 & 26.86 \\
         RoBERTa$_{\text{COSMOS}}$ & 25.14 & 28.73 \\
         \midrule
         RoBERTa$_{\text{RACE} \longrightarrow \text{\dataname}}$ & 34.60 & 35.07 \\
         RoBERTa$_{\text{COSMOS} \longrightarrow \text{\dataname}}$ & 36.44 & 35.11                                                   \\
         \bottomrule
    \end{tabular}
    \caption{Transfer learning results (accuracy\%).}
    \label{tab:transfer_learning}
\end{table}

Ablating question causes a relatively smaller performance drop as compared with the paragraph, which is consistent with observations by Huang \emph{et  al.}\shortcite{cosmos}. This is likely because the diversity of questions is lower.
The above results show that our dataset does not have a strong bias.

\begin{table}[t!]
    \centering
    \setlength{\abovecaptionskip}{0.1cm}
    \setlength{\belowcaptionskip}{-0.5cm}
    \scriptsize
    \begin{tabular}{c|cccc}
        \toprule
         {\bf Length} & {\bf (0,100]} & {\bf (100,150]} & {\bf (150,200]} &{\bf (200,$+\infty$)}\\
         \midrule
         \#Instances & 253 & 364 & 198 & 61 \\
         RoBERTa & 31.31 & 36.25 & 37.13 & 40.38\\
         \bottomrule
    \end{tabular}
    \caption{Performance of different
length (accuracy\%).}
    \label{tab:length performance}
\end{table}

\subsection{Transfer Learning}
We conduct a set of transfer learning experiments to understand the degree of overlap in terms of necessary knowledge for solving problems in our dataset and existing datasets. In particular, we first fine-tune the RoBERTa model on a source dataset, before fine-tuning the model on \dataname. If the required knowledge is similar, the model performance is expected to increase. RACE and COSMOS are adopted as the source datasets. The former tests English reading skills while the latter tests commonsense knowledge. As shown in Table~\ref{tab:transfer_learning}, the RoBERTa model trained only on either source dataset gives significantly lower accuracies on \dataname \ test set compared with the RoBERTa model trained on \dataname. The performance of RoBERTa trained on RACE is even close to the random guess baseline. In addition, further fine-tuning on \dataname\ leads to improvements over the source-trained baselines, but the resulting models do not outperform a model trained only on \dataname. The observation is different from most other datasets \cite{cosmos,Social-IQa}, which demonstrates that \dataname \ contains highly different challenges compared with existing datasets.

\subsection{Performance Across Different Lengths}
We measure the accuracy of RoBERTa against the input size. In particular, the number of words in the paragraph, the question and the candidate answers are added together as the length of a test instance. The statistics and performances are all shown in Table~\ref{tab:length performance}. Interestingly, the model performances are not negatively associated with the input size, which is different from most NLP benchmarks. This shows that the level of challenge in logical reasoning can be independent of the input verbosity.

\subsection{Lexical Overlap}
We aim to understand a bias of models in selecting the candidate answers that have the best surface matching with the paragraph. To this end, we calculate the unigram overlap between each candidate answer and the given paragraph for each problem, and mark the best-matching candidate. 
We report the ``Overlap Ratio'' by calculating the accuracy between model prediction and the best-matching candidate.
The results are shown in Table~\ref{tab:model_reasoning_type}. As can be seen, the gold-standard output has an accuracy of 28.37\%, whilst all of the models give accuracies above this number, which shows a tendency of superficial matching. In particular, the word matching method gives an accuracy of 100\% due to its mechanism. RoBERTa gives the lowest matching accuracy, showing that it relies the least on lexical patterns. We additionally measure the accuracy of the models on the ``correct'' instances according to best matching. RoBERTa still outperforms the other models, demonstrating relative strength in logical reasoning.

\begin{figure}[t!]
\centering
\setlength{\abovecaptionskip}{0.cm}
\setlength{\belowcaptionskip}{-0.3cm}
\includegraphics[width=0.48\textwidth]{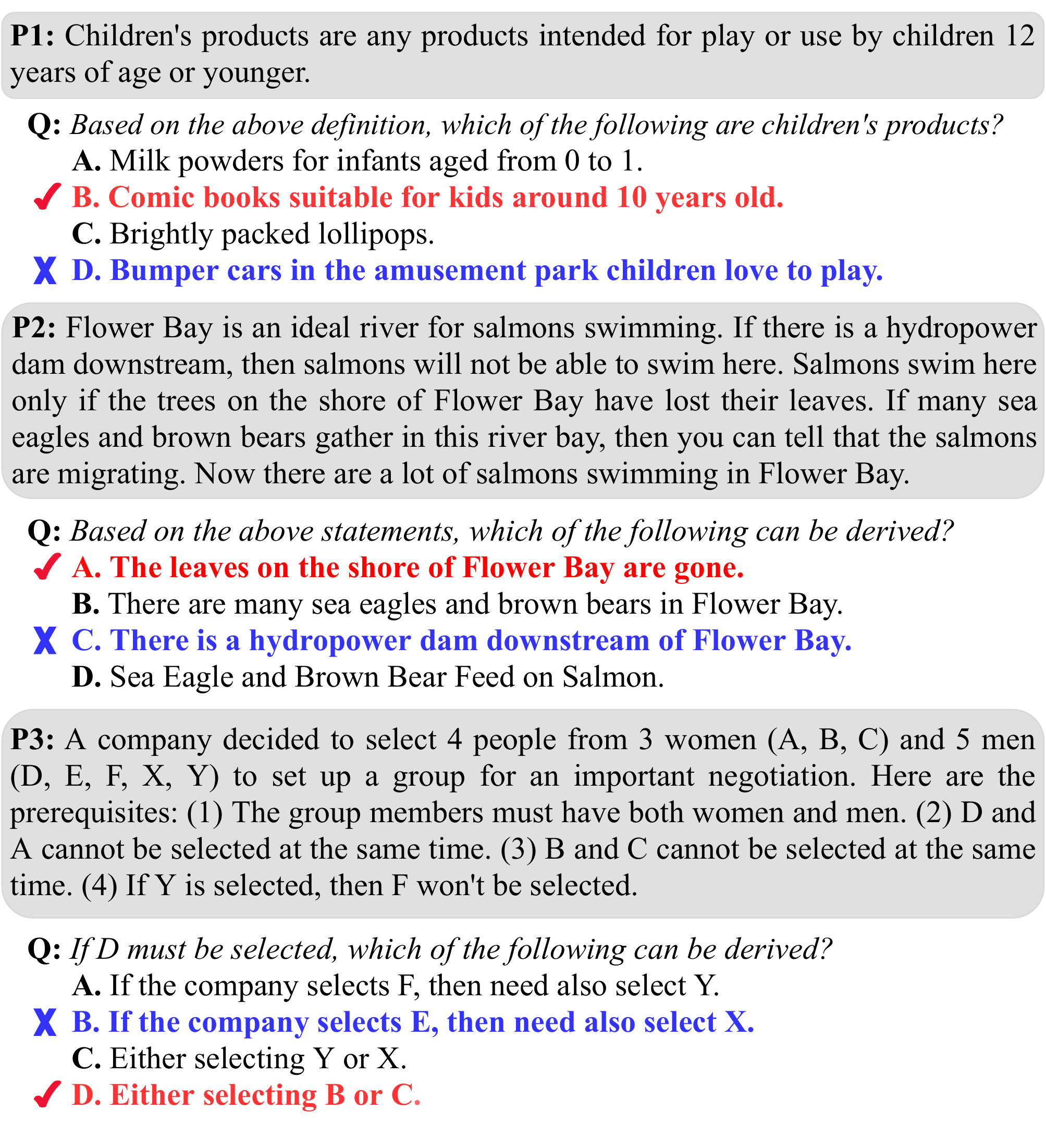}
\caption{Example mistakes of RoBEETa. 
({\color{red} {\cmark}} indicates the correct answers and {\color{blue} {\xmark}} indicates the RoBERTa prediction.)}
\label{fig:error}
\end{figure}

\subsection{Reasoning Types}
Table~\ref{tab:acc_reasoning_type} gives the performances of RoBERTa over the 5 reasoning types discussed in Section~\ref{sec:reasoning type}. The method gives the best accuracy on categorical reasoning. However, for the other four reasoning types, the results are significantly lower. To understand why these tasks are challenging for RoBERTa, we give qualitative discussion via case study.

\paragraph{Categorical reasoning:} P1 of Figure~\ref{fig:error} shows a typical example, where the definition of children's products is given in the paragraph and the testee is asked to select a correct instance. A key here is the age range (i.e., under 12). RoBERTa incorrectly chooses the candidate that is superficially similar to the paragraph, while ignoring the reasoning process.



\paragraph{Conditional reasoning:} 
P2 of Figure~\ref{fig:error} is a representative example of the most challenging conditional reasoning questions. In particular, a variety of sufficient and necessary conditional relations are given in the paragraph, which include:
\begin{equation*}
\small
\begin{split}
&x = ``\text{Salmons\ swim}" \\
&y = ``\text{Sea\ eagles\ and\ brown\ bears\ gather}" \\
&z = ``\text{Hydropower\ dam\ exists\ downstream}" \\
&w = ``\text{Trees\ lose\ leaves}" \\
&x => w \ \text{(Necessary\ conditional\ relation)}\\
&y => x \ \text{ (Sufficient\ conditional\ relation)}\\
&x => \overline{z} \ \text{ (Sufficient\ conditional\ relation)}
\end{split}
\end{equation*}
\quad The correct answer depends on fully understanding both the necessary and sufficient conditional reasoning facts. RoBERTa makes a mistake by ignoring the ``\emph{not}" operator in the $x => \bar{z}$ condition, which coincides with prior observations on BERT and negation \cite{niven-kao-2019-probing}.




\paragraph{Conjunctive and disjunctive reasoning:} 


\begin{table}[!t]
\centering
\setlength{\abovecaptionskip}{0.1cm}
\setlength{\belowcaptionskip}{-0.3cm}
\scriptsize
\begin{tabular}{l|c|c}
\toprule
{\bf Model} & {\bf Overlap Ratio} & {\bf Accuracy(\%)} \\ 
\midrule
Word Matching & 100.00 & 28.37 \\
Stanford Attentive Reader & 35.47 & 35.82  \\
Gated-Attention Reader & 37.33  & 34.96 \\ 
Co-Matching Network & 40.74 & 36.85 \\
BERT & 34.23 & 37.73 \\ 
RoBERTa & 32.08 & 40.38 \\ 
\midrule
Gold-standard & 28.37 & 100.00\\
\bottomrule
\end{tabular}
    \caption{Overlap ratio (\%) against the model type.}
\label{tab:model_reasoning_type}
\end{table}

\begin{table}[!t]
\centering
\setlength{\abovecaptionskip}{0.1cm}
\setlength{\belowcaptionskip}{-0.3cm}
\scriptsize
\begin{tabular}{l|cc}
\toprule
{\bf Reasoning Type} & {\bf RoBERTa} \\ 
\midrule
Categorical reasoning & 55.00 \\
Sufficient conditional reasoning & 17.11 \\
Necessary conditional reasoning & 19.29 \\
Disjunctive reasoning & 22.67 \\ 
Conjunctive reasoning & 21.98 \\ 
\bottomrule
\end{tabular}
    \caption{RoBERTa accuracy (\%) against the reasoning type.}
\label{tab:acc_reasoning_type}
\end{table}

P3 of Figure~\ref{fig:error} represents one of the most challenging questions in the dataset, where the premises and candidate give a set of constraints in both conjunctive and disjunctive forms, and the question asks which candidate conforms to the premises. The testee is expected to enumerate different possible situations and then match the cases to the candidates by thoroughly understanding the candidates also. Intuitively, RoBERTa is not directly equipped with such reasoning capacity.

\section{Conclusion}
We have presented \dataname, a large-scale logical reasoning reading comprehension dataset. In addition to testing reasoning capacities of machine reading, our dataset can also serve as a benchmark for re-examining the long pursued research of logical AI in the deep learning NLP era. Results show that the state-of-the-art machine readers still fall far behind human performance, making our dataset one of the most challenging test for reading comprehension.

\section*{Acknowledgments}

This work is supported by the National Science Foundation of China (Grant No. 61976180). We also acknowledge funding support from the Westlake University and Bright Dream Joint Institute for Intelligent Robotics.


\bibliographystyle{named}
\bibliography{ijcai}

\end{document}